\documentclass[letterpaper, 10 pt, conference]{ieeeconf}  

\IEEEoverridecommandlockouts                              

\usepackage{graphics} 
\usepackage{multirow}
\usepackage{graphicx}
\usepackage{amsmath}
\usepackage{amssymb}
\usepackage{cite} 
\usepackage{soul} 
\usepackage{xcolor}
\sethlcolor{yellow}

\makeatletter
\renewcommand{\@listI}{%
  \leftmargin=2em
  \itemindent=0pt
  \parsep=0pt
  \topsep=0pt
  \itemsep=0pt
  \partopsep=0pt
}
\makeatother

\makeatletter
\def\endfigure{\end@float}
\def\endtable{\end@float}
\makeatother

\usepackage[font=normalsize,
            labelfont={bf},
            labelsep=period,
            justification=centering,
            singlelinecheck=false]{caption}
\DeclareCaptionFormat{ieee}{\centering#1#2#3}
\usepackage{subfig}

\usepackage{cleveref} 
\usepackage{cancel} 

\title{\LARGE \bf
CAFEs: Cable-driven Collaborative Floating End-Effectors for Agriculture Applications}

\author{
\thanks{*This work was supported by EPFL Solutions4Sustinability Grant}
\thanks{Note: This work has been submitted to the IEEE for possible publication. Copyright may be transferred without notice, after which this version may no longer be accessible.}%
Hung Hon Cheng$^{1}$ and Josie Hughes$^{2}$
\thanks{$^{1}$ Hung Hon Cheng is with the CREATE Lab, School of Engineering STI, EPFL, Swiss
        {\tt\small hung.cheng@epfl.ch}}%
\thanks{$^{2}$Josie Hughes is with the CREATE Lab, School of Engineering STI, EPFL, Swiss {\tt\small josie.hughes@epfl.ch}}%
}

\newcommand{\atan}{\operatorname{atan}}

\begin{document}
\maketitle
\thispagestyle{empty}
\pagestyle{empty}

\begin{abstract}

\emph{CAFEs} (Collaborative Agricultural Floating End-effectors) is a new robot design and control approach to automating large-scale agricultural tasks. Based upon a cable driven robot architecture, by sharing the same roller-driven cable set with modular robotic arms, a fast-switching clamping mechanism allows each \emph{CAFE} to clamp onto or release from the moving cables, enabling both independent and synchronized movement across the workspace. The methods developed to enable this system include the mechanical design, precise position control and a dynamic model for the spring-mass liked system, ensuring accurate and stable movement of the robotic arms. The system's scalability is further explored by studying the tension and sag in the cables to maintain performance as more robotic arms are deployed. Experimental and  simulation results demonstrate the system's effectiveness in tasks including pick-and-place showing its potential to contribute to agricultural automation.

\end{abstract}

\section{INTRODUCTION}
Robotic systems in agriculture primarily rely on Unmanned Ground Vehicles (UGVs) for harvesting \cite{polzin2023automation,birrell2020field,xiong2020autonomous} and Unmanned Aerial Vehicles (UAVs) for monitoring and yield estimation \cite{turner2011development,kim2019unmanned,xiang2011development,torres2014multi,sarri2019testing}. While these systems effectively navigate complex farm environments, they rely on GPS, Real-Time Kinematic (RTK) positioning, Simultaneous Localization and Mapping (SLAM), or visual-inertial odometry for self-localization \cite{oliveira2021advances}. However, these localization methods can be intricate, adding to the challenges of cost, scalability, and coordination for large-scale deployment. A well-designed cable-driven mechanism could offer a viable alternative, reducing localization dependency while maintaining efficient coverage of agricultural fields.

Cable-driven robots offer an alternative approach by providing precise positioning without complex navigation \cite{garcia2023agrocablebot,tai2019design,prabha2021cable}. Examples like SkyCam \cite{tanaka1988kineto} and CU-brick \cite{wu2018cu} showcase their advantages in large spaces without the need for complex navigation, making them useful for
agricultural work. However, traditional cable-driven robots are constrained by their scalability due to the limited number of end-effectors.The multi-robot coordination is challenging regarding to the cable interference \cite{cheng2022ray}. Although reconfigurable cable robot \cite{jamshidifar2020static,cheng2023cable,xiong2022real} improve workspace flexibility, they introduce hardware and control complexities due to multiple interacting cables.

Suspended cable-driven systems can avoid some of these limitations, but they often suffer from limited workspaces near the ground \cite{saber2013workspace} and require complex real-time control to mitigate oscillations \cite{alp2002cable,korayem2010dynamic,alikhani2011modeling}. Additionally, when scaling up these systems to tens or hundreds of meters, cable elongation and flexing under variable loads can introduce positioning errors in the centimeter range, while differential sagging affects precision, limiting their use in large-scale farming.

\begin{figure}[t] \centering \includegraphics[width = 0.45\textwidth]{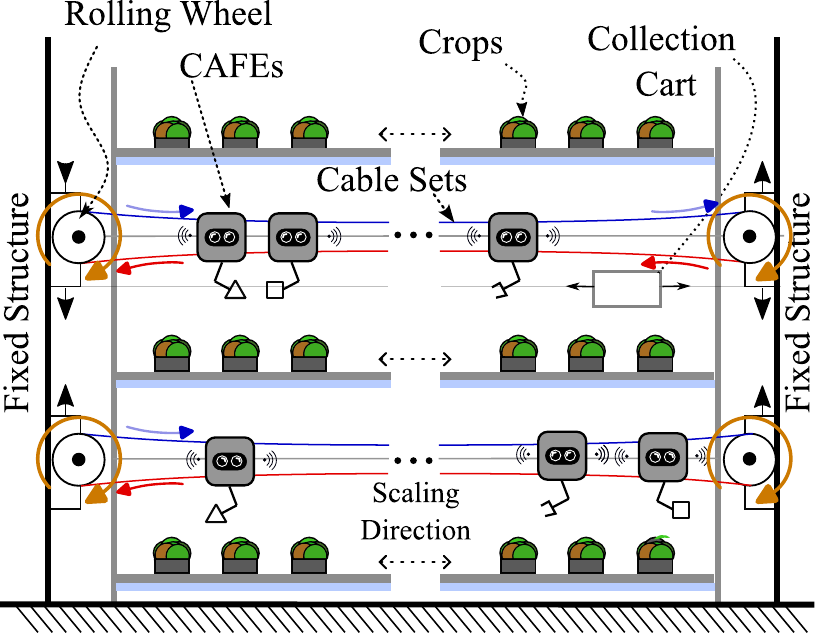} \caption{Application scene overview in a greenhouse and vertical farming} \label{fig:fisrt_figure} \end{figure}

The challenges of using mobile and cable-driven systems highlight the need for a solution that can cover large areas
and allow multiple robots to work together efficiently. We propose a cable-driven system for collaborative end effectors
(CAFEs) that adresses these problems by introducing a cable-driven setup that allows several robotic arms to work
together without requiring complex navigation or workspace rearrangement as shown in \Cref{fig:fisrt_figure}.
The proposed system leverages the advantages of cable-driven robots while enabling cooperation between multiple end-effectors, such as for harvesting tasks. It can be deployed in both greenhouses and outdoor fields by strategically placing multiple anchoring points. The system's scalability is achieved by adding more CAFEs, incorporating additional cable sets, or extending the cable lengths.

The proposed cable-driven system features both passive cables and a commonly shared roller-driven cable that runs along the length of the floating structure. By providing each floating platform with the ability
to independently clamp on and off (in either direction of the cable) we maintain the potential for independent operation
whilst enabling far simpler synchronzied control as they can be moved by the same rollder-driven cable. Because the design allow multiple robots traversing on the the set of cable, such terrain-independent but still coordinated advantage is able to address the problem in both UGVs and UAVs. Also, the robotic arms can travel in the air with low energy consumption due to the shared weight on the cables. Furthermore, asthe actuation is provided by the shared moving cable each platform remains light weight and requires only clamping control. Specifically the contributions of this work include:

\begin{enumerate} 
\item A novel design of a cable traversing system that enables both individual and synchronized movement of multiple robots with simple logic control.

\item Demonstration of the precise and repeatable position control of the system.
\item Use of a dynamic model to analyze the tension distribution and scalability of the robot.
\end{enumerate}

The subsequent section of this paper  detail the robots design, control and dynamic behaviour. The experimental setup and results are then presented, finishing with a conclusion and discussion.

\begin{figure}[h]
    \centering    \includegraphics[width=0.95\linewidth]{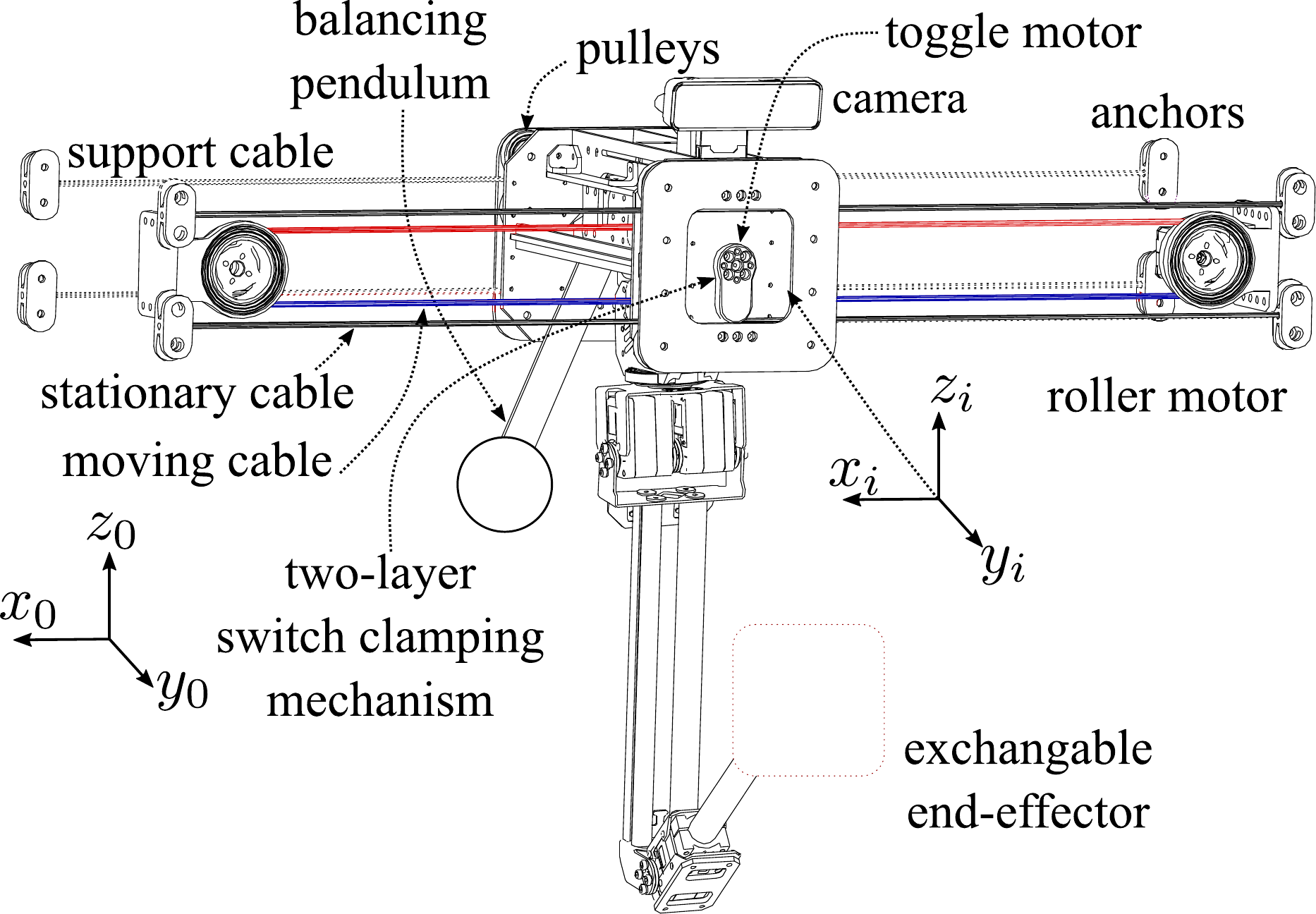}
    \caption{Mechanical design of individual \emph{CAFE}}
    \label{fig:overview_ccafe}
\end{figure}

\section{METHODS}

The proposed system (\Cref{fig:fisrt_figure}) comprises multiple \emph{CAFEs} mounted on a shared three-cable setup, with two stationary passive cables and one actively driven cable for movement. Each \emph{CAFE} (\Cref{fig:overview_ccafe}) integrates a robotic arm, clamping mechanism, and camera for harvesting, powered by an exchangeable LiPo battery that supplies the Raspberry Pi and servo motors. Wireless ROS2 communication ensures coordinated control. The cable set is anchored to a rigid structure and pre-tensioned based on cable sag, \emph{CAFEs} count, and total length. The following section details the system’s design and control.

\subsection{Cable System and Operation Mode}

The cable structure of the system has three main sets of cables, each serving a different purpose:
\begin{itemize}
\item Continuously moving cable loop (red and blue color in \Cref{fig:overview_ccafe}). This is driven by two roller motors, rotating at the same direction and providing both leftward (lower cable) and rightward (upper cable) motion to any attached \emph{CAFEs}. 
\item Stationary cable sets. Pretensioned and anchored at the cable ends, acting as motion stoppers for the robot and support the weight. 
\item Load-supporting cable sets:  Pretensioned and anchored at the cable ends, designed to support the robot's weight only. \end{itemize}

All \emph{CAFEs} in the system share the three cable sets, reducing the need for extra components and powerful motors on the platform. This design has two advantages. First, it minimizes both cable sag and pretension caused by the accumulated robots' weight.  Secondly, it ensures synchronized movement for precise position control, avoiding slippage issues which are common in traditional cable-traversing robots with rolling wheels. The use of two roller motors guarantees sufficient torque to move multiple \emph{CAFEs} efficiently along  long cables.

\begin{figure}[h]
    \centering
    \begin{minipage}[b]{0.4\textwidth}
        \centering
        \includegraphics[width=\textwidth]{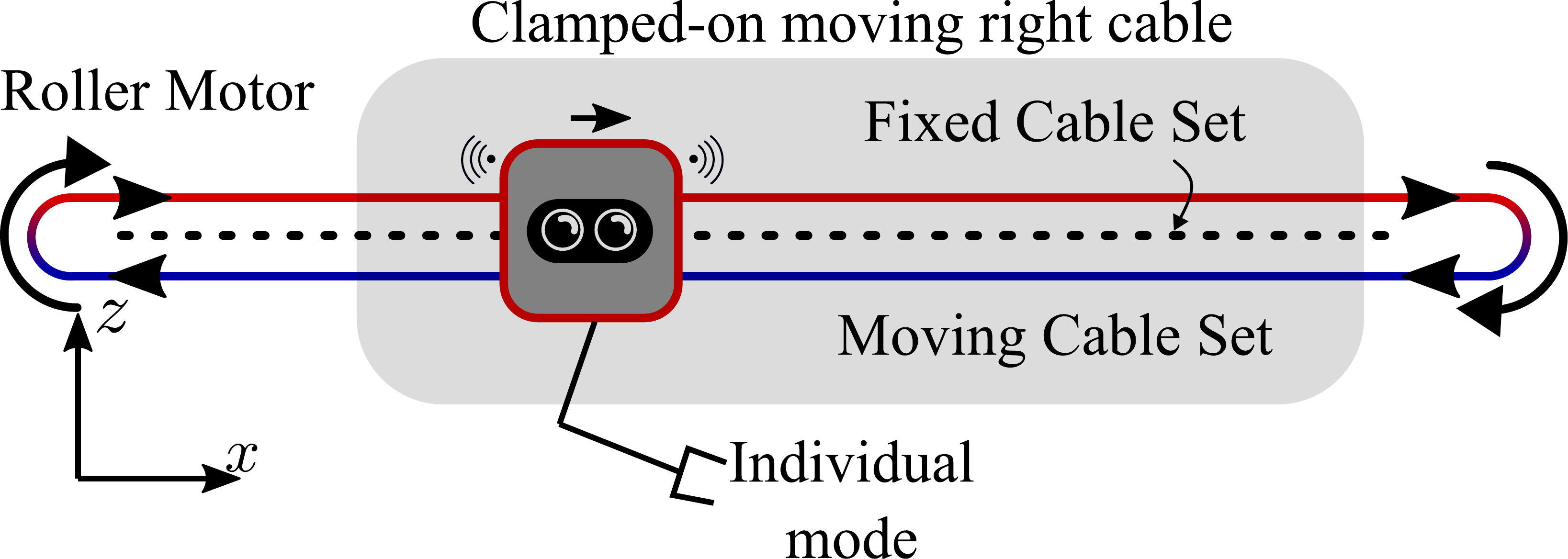}
        \centerline{(a) Individual CAFE}
    \end{minipage}
    \begin{minipage}[b]{0.4\textwidth}
        \centering
        \includegraphics[width=\textwidth]{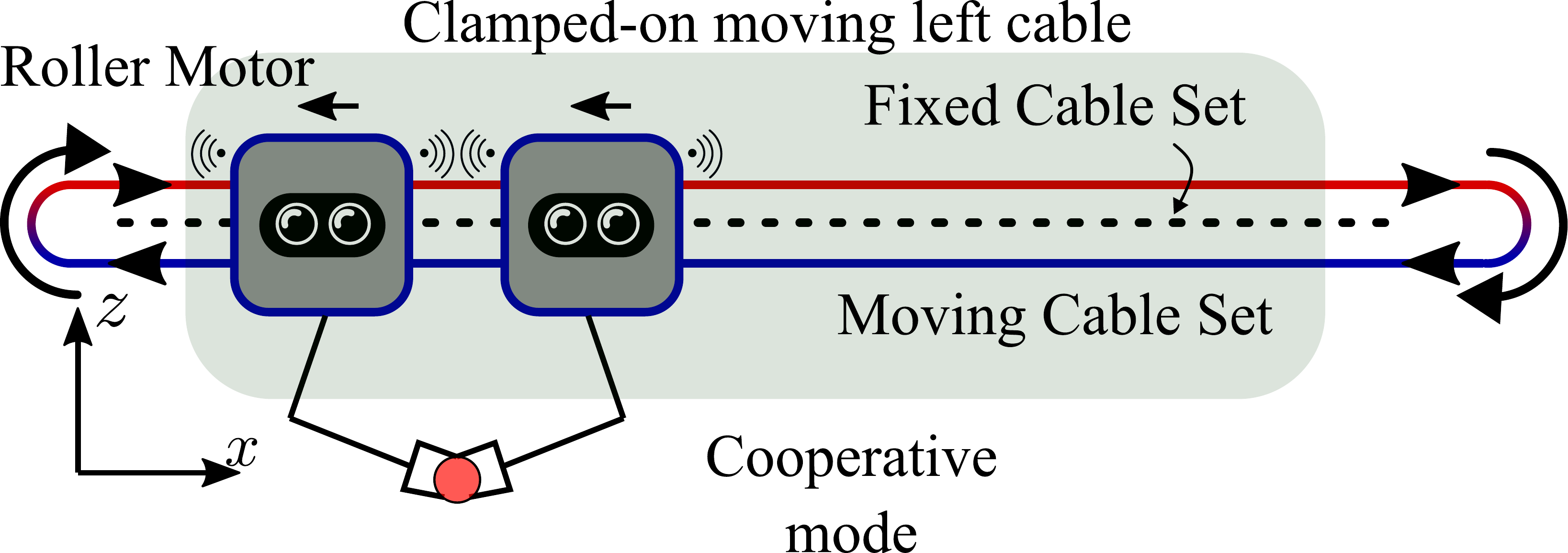}
        \centerline{(b) Cooperative between CAFEs}
        \label{fig:indiviual_mode}
    \end{minipage}
    \begin{minipage}[b]{0.4\textwidth}
        \centering
        \includegraphics[width=\textwidth]{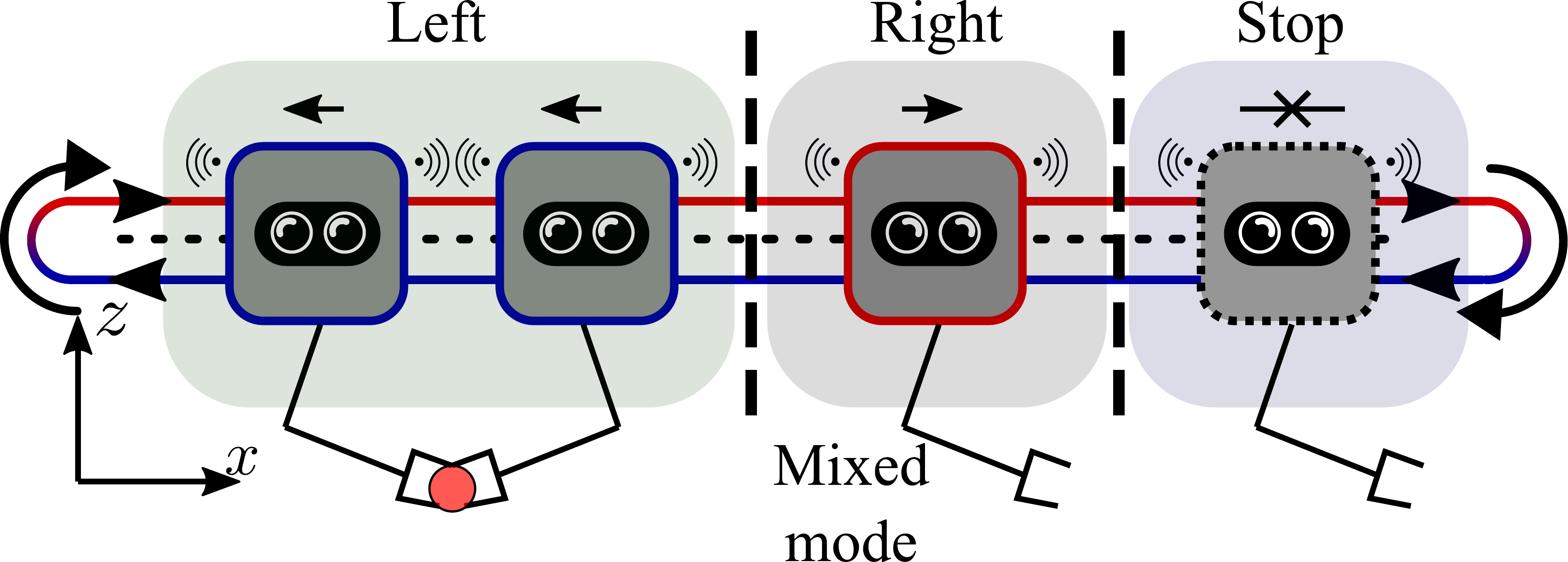}
        \centerline{(c) Mixed operation mode}
        \label{fig:cooperative_mode}
    \end{minipage}
    \caption{Different operation mode for the CAFEs system}
    \label{fig:CAFEs_system}
\end{figure}

Each of the CAFEs can anchor on to the continuously moving cable or stationary cable. This enables a number of modes of operation for a multi-robot system which are summarized in \Cref{fig:CAFEs_system}:  
\begin{itemize} 
\item Indepedent Individual Mode. Each single lightweight \emph{CAFE} can clamp on/off to the actuated/stationary cable having independent position control in either direction.  This is useful when each arm is perform a different tasks, for example, crop inspection and monitoring in farms. 
\item Cooperative Mode. When two or more \emph{CAFEs} need to move a fixed distance together  (in symmetry or anti-symmetry), by clamping on/off at the same time their separation distance can be guaranteed as they are being moved by the same driving cable.  This is useful when the arms are performing bi-manual lifting, or a mirrored task.
\item Mixed Mode: Different \emph{CAFEs} can operate in different modes along the cable system. We make the assumption that each robot, or set of robots will have a fixed area of operation.

\end{itemize} 

\subsection{Switch Clamping Mechanism}

\begin{figure}[!htb]
    \centering    \includegraphics[width=0.75\linewidth]{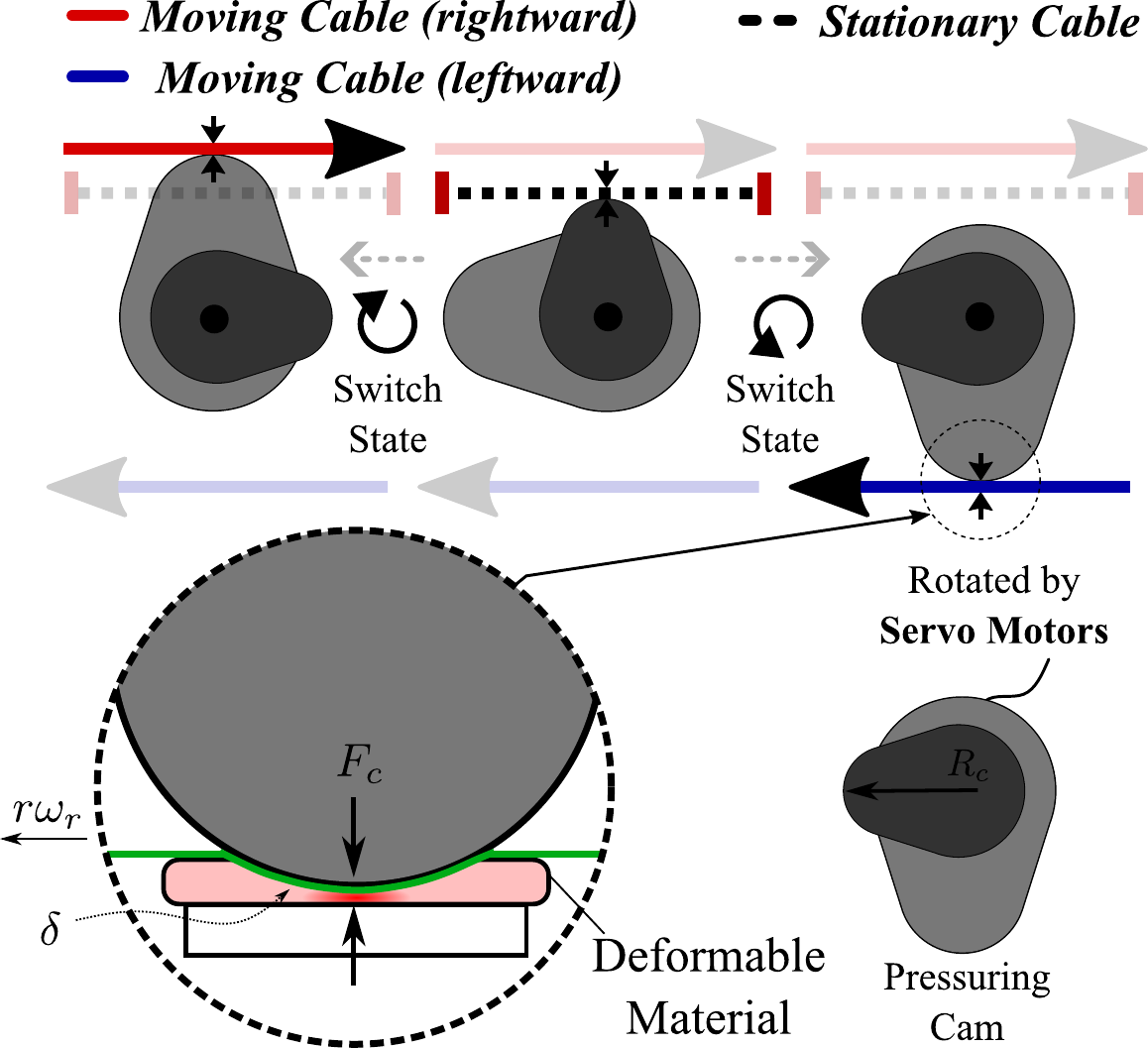}
    \caption{Two-layer switch clamping mechanism and the three motion states corresponds to the moving direction}
    \label{fig:shematic_clamping}
\end{figure}

As shown in \Cref{fig:shematic_clamping}, the roller motors and continuous cable provides two moving directions of the cables, and the stationary cable a third possible state. Thus, the robot requires a clamping mechanism which can enable anchoring to one of these there cables. This is performed with a pressuring cam (illustrated in \Cref{fig:shematic_clamping}) which rotates $90^\circ$ to contacts the soft materials fixed on the robot. The clamping angles are $\theta_{l}$, $\theta_{r}$, and $\theta_{s}$ for three moving states: leftward, rightward, and stationary, respectively. The pressuring cam 
generating a normal force $F_{c}$, which follows a $\frac{3}{2}$ relationship with material deformation $\delta$, according to the Hertzian contact force equation to the elastic half-space:

\begin{equation}
F_{c} = \frac{4}{3} E \sqrt{R_c} \delta^{3/2} \nonumber
\end{equation}
where $R_c$ represents the radius of the pressuring cam, $\delta$ denotes the total deformation, and $E$ stands for the Young's modulus ratio. Thus, the clamping force produces high friction, even without high cable tension, to anchor the robot on to the required cable. As such the robot can be securely hold at its position when needed without any additional control of energy.

\subsection{Floating Platform and position control}

Each \emph{CAFE} integrates components such as pulleys, a toggle motor with the clamping mechanism, a depth camera, and a counterbalancing system. The depth camera provides feedback for navigating toward the target region. In the next development, the depth camera is also used to deploy the vision tracking to locate the crops with the YOLO or other approaches.
The counterbalancing system stabilizes the platform in response to the robotic arm's movements.
Driven by the roller motors with the same angular speed $\omega_r$ and a radius $r$, the position of the $i^{th}$ \emph{CAFE} is determined by
\begin{equation}
\begin{aligned}
\left [
\begin{matrix}
x_{i}(t) \\
z_{i}(t)
\end{matrix}
\right]
= 
\left[
\begin{matrix}
s\Delta D(t) \cos(\theta_{I}) + x_{i0} \\
s\Delta D(t) \sin(\theta_{I}) + z_{i0}
\end{matrix}
\right]
\end{aligned}
\label{equ:pos_of_floating_platform}
\end{equation}
where $[x_{i0}, z_{i0}]^T$, $s\in \left[ -1,0,1\right]$, $\Delta D$ and $\cos(\theta_{I})$ is the previous state's position, moving direction, linear displacement of driving cable, and incline angle to horizontal, respectively. In addition, the linear displacement of driving cable $\Delta D$ is calculated by:

\begin{equation}
    \label{equ:min_distance_change}
    \Delta D(t) = r\omega_r (t_a - t_s) 
\end{equation}

where $t_s$ and $t_a$ is clamping-on and clamping-off moment. $\omega_r$ is the roller angular velocity and $r$ denotes the roller's radius. Precise positioning of the floating platform is achieved through simple logic control of the clamping timing.

\subsection{Dynamic System Model \& Behaviour}\label{sec_Ten}

The system can be modeled as a spring-mass system, where the horizontal cables act as springs and the \emph{CAFEs} as masses. Cable sag affects the robot’s vertical position during traversal, necessitating an analysis of sag and oscillation dynamics for effective active position compensation, particularly in multi-\emph{CAFE} setups.
\begin{figure}[h]
    \centering    \includegraphics[width=0.72\linewidth]{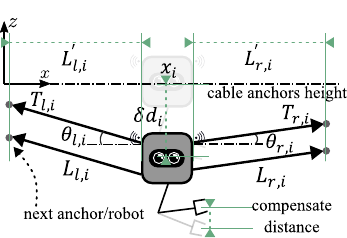}
    \caption{Tension and elongation due to the robot's mass on the cable system}
    \label{fig:load_disturbution}
\end{figure}
\begin{figure*}[!t]
    \centering   
    \subfloat[Response with different loads]{\label{fig:motion_1}
    \includegraphics[width = 0.21\textwidth]{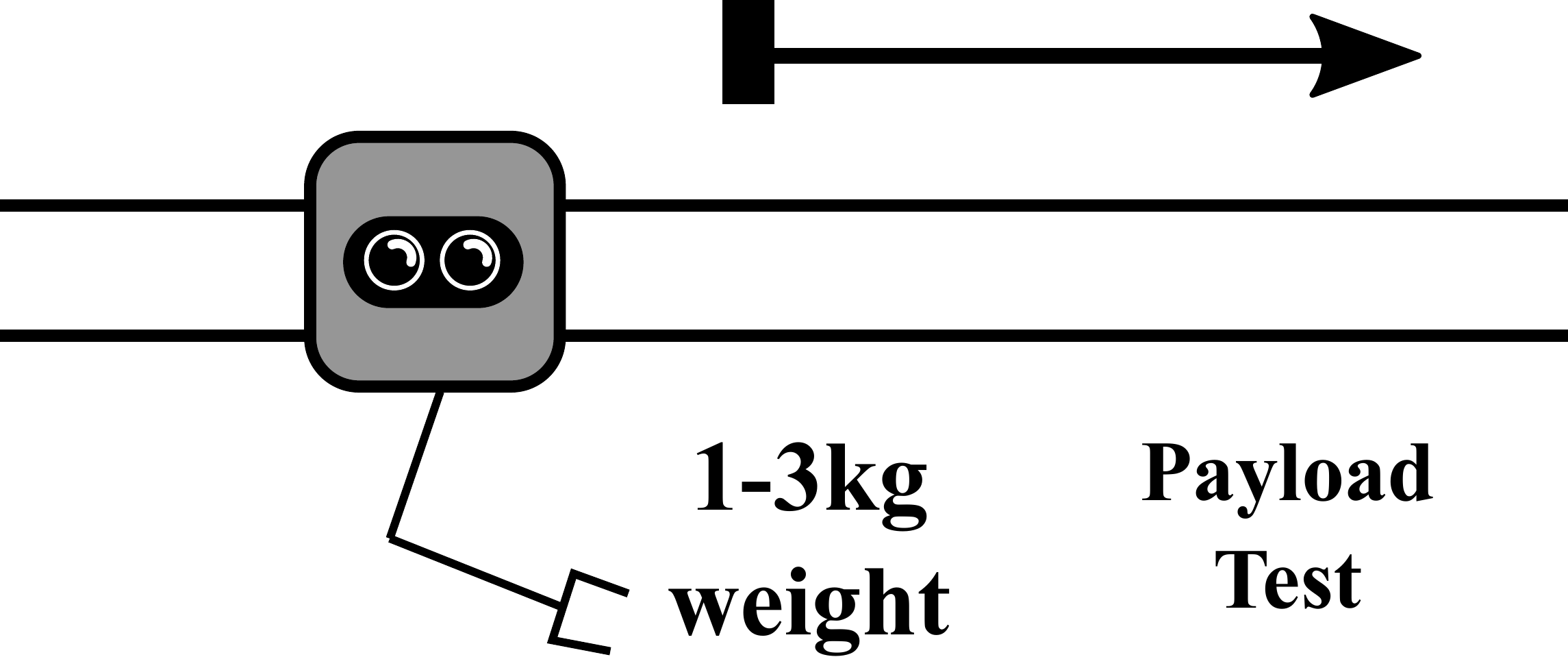}}
    \subfloat[Intermittent motion]{\label{fig:motion_2}
    \includegraphics[width = 0.23\textwidth]{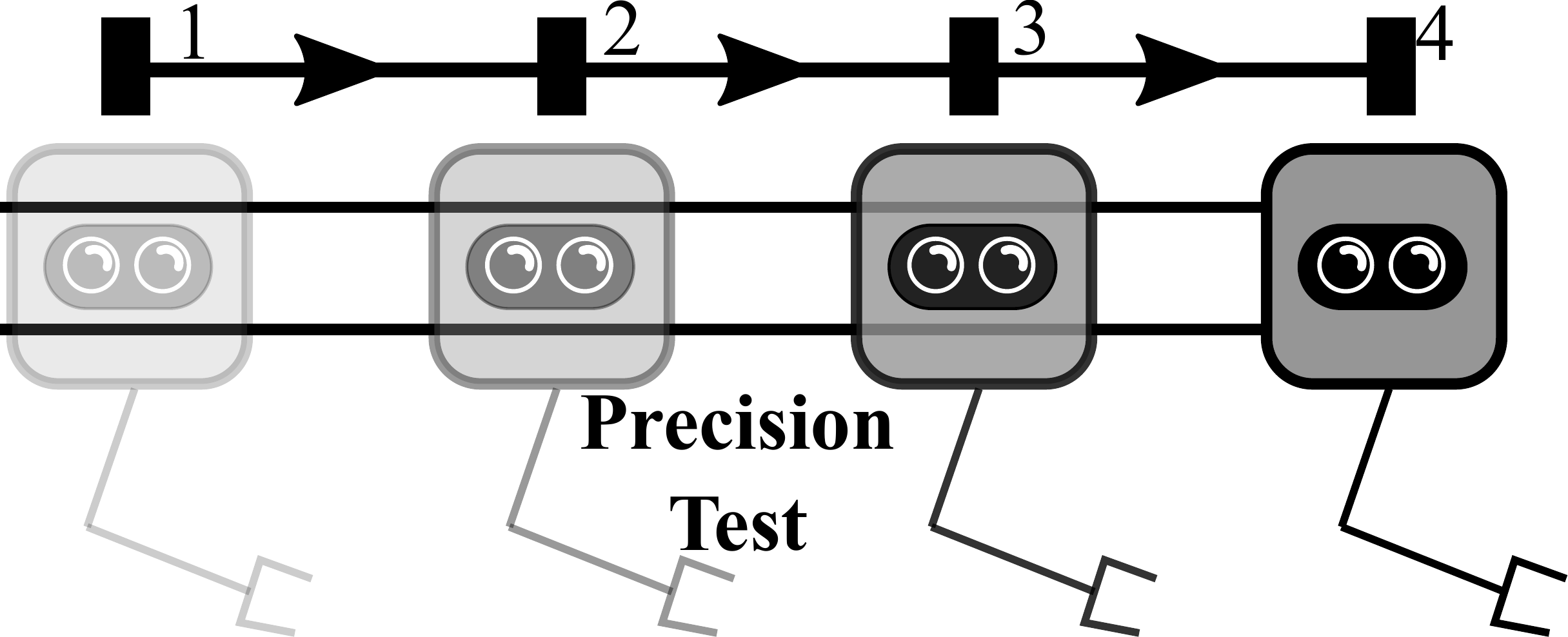}}
    \subfloat[Cyclic between two points]{\label{fig:motion_3}
    \includegraphics[width = 0.22\textwidth]{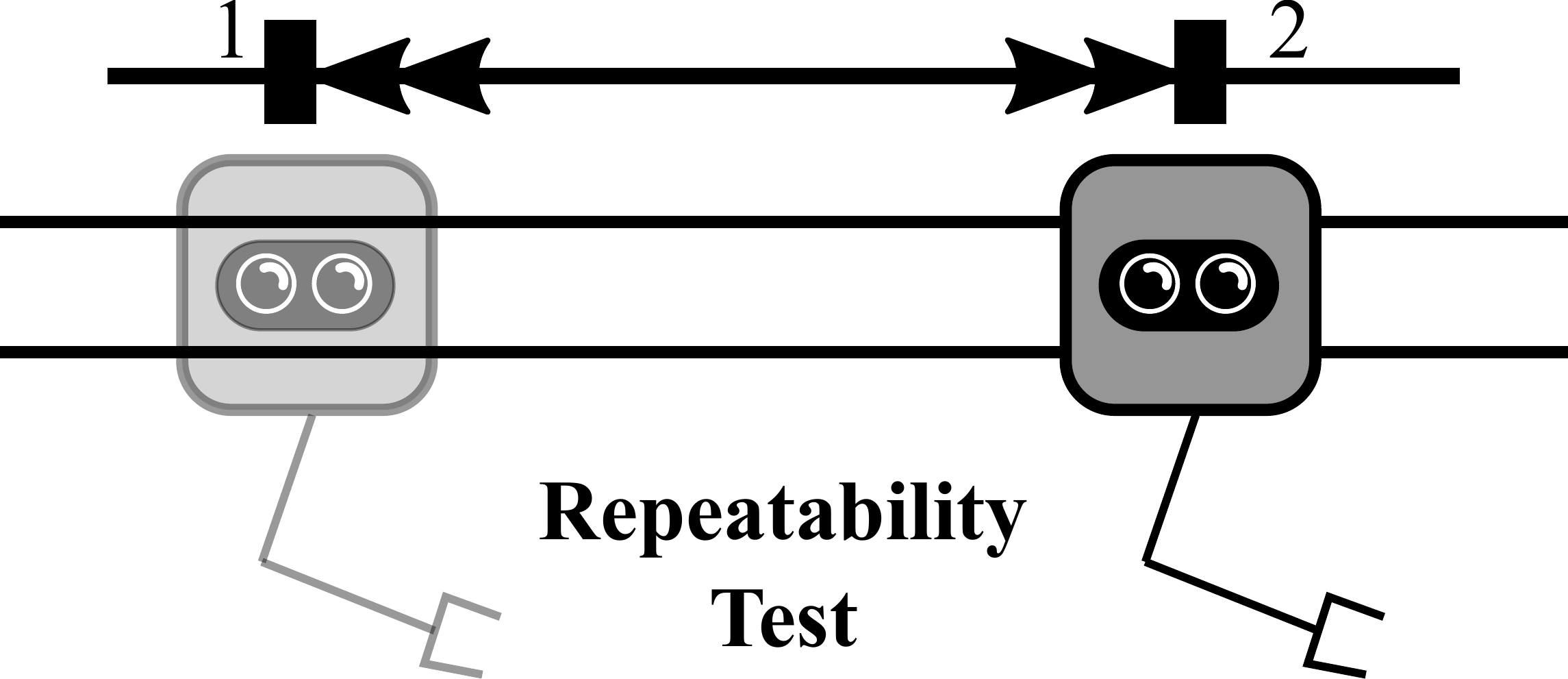}}
    \subfloat[Synchronous motion]{\label{fig:motion_5}
    \includegraphics[width = 0.22\textwidth]{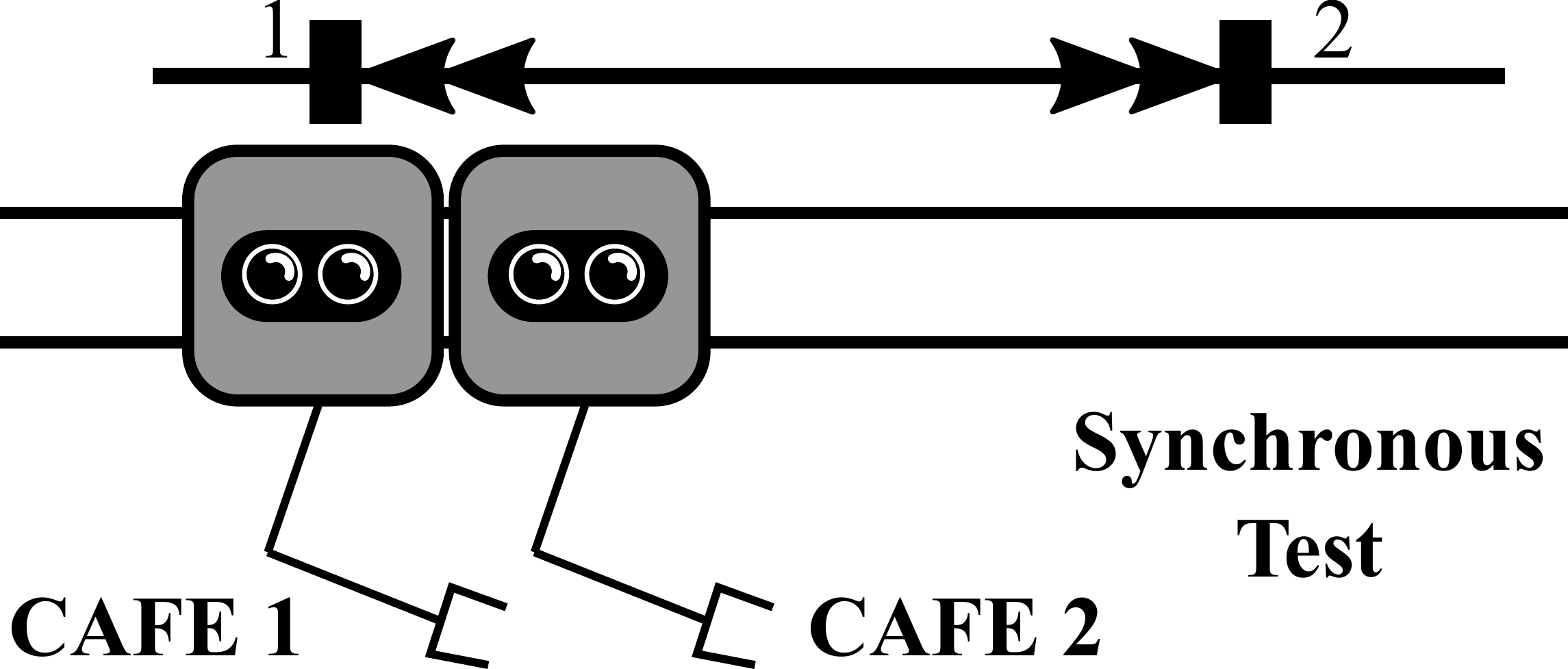}}
    \caption{Motions for different experiments}
    \label{fig:motion_fig_set}
\end{figure*}

As shown in \Cref{fig:load_disturbution}, the $i^{th}$ \emph{CAFE} will experience a vertical displacement of $\delta d_i$.  The equation of motion for each individual \emph{CAFE} can be expressed as
\begin{equation} \label{equationofmotion}
\begin{gathered}
m_i \ddot{z}_i = T_{l,i} \sin(\theta_{l,i}) + T_{r,i} \sin(\theta_{r,i}) - m_i g - c \dot{z}_i \\
T_{l,i} \cos(\theta_{l,i}) = T_{r,i} \cos(\theta_{r,i})
\end{gathered}
\end{equation}
where $\ddot{z}_i$, $\dot{z}_i$, $T_{l,i}$, $T_{r,i}$, $\theta_{l,i}$, $\theta_{r,i}$, $m_i$, $g$, and $c$ represent the acceleration, velocity, tensions, angles, mass, gravitational constant, and damping coefficient, respectively. Specifically, the angles are
\begin{equation}
    \theta_{l,i} = \atan\left(\frac{z_{i-1}-z_i}{x_{i-1}-x_i}\right),  \theta_{r,i} = \atan\left(\frac{z_{i+1} - z_i}{x_{i+1} - x_i}\right) \nonumber
\end{equation}
and tension values on each cable segments are calculated by
\begin{align}
T_{l,i} &= \textit{abs}(k \cdot \max(0, L_{l,i} - L_{l,i}^\prime)) + \textit{abs}(T_{pre}) \nonumber\\
T_{r,i} &= \textit{abs}(k \cdot \max(0, L_{r,i} - L_{r,i}^\prime)) + \textit{abs}(T_{pre}) \nonumber
\end{align}
where $T_{pre}$, $L_{l,i}$, $L_{r,i}$, $L_{l,i}^\prime$ and $L_{r,i}^\prime$ are pretension, the $i^{th}$ current cable lengths and original cable lengths to the next anchor or robot, respectively. Since the cable cannot be compressed, the maximum value between the elongation and zero, such as $\max(0, L_{l,i} - L_{l,i}^\prime)$, should be considered. Also, $k$ is the stiffness of the cable, which assumed all the cables have same stiffness for simplicity. It is worth to notice that the sign of the tension force should be determined by the vertical position of the robot such that
\begin{equation}
T_{l,i} = -T_{l,i}, \ \text{if} \ z_i > z_{i-1}, \quad T_{r,i} = -T_{r,i}, \ \text{if} \ z_i > z_{i+1} \nonumber
\end{equation}
Using the above equations, the tension distribution and the vertical displacement of each robot can be determined by solving the ODE equations. As such, the vertical displacement of $\delta d_i$ can be estimated and compensated by adding into the $\mathbf{x}_{i}(t)$ of the robotic arm. Additionally, the oscillation behaviours can also be predicted and controlled by adding the damping system,

\subsection{Scalability Analysis for Multiple Floating Platforms}
Scaling the system with multiple CAFEs requires managing cable load limits and dynamic interactions. Increased pretension reduces sag, with required values estimated from previous equations and validated in the results. For short-range (2–5m) deployments in small greenhouses, lower pretension (20–100kg) is sufficient due to fewer arms. Medium-range and full-span deployments, such as open-field applications, demand higher pretension, which can be mitigated by reducing arm weight and employing active tension monitoring, distributed anchor points (every 5m), and real-time position compensation using a spring-mass model.

Each CAFE expands the workspace by traversing the cable network, with its reach defined by kinematics (\Cref{fig:indiviual_mode}). Multiple CAFEs enable a scalable, collaborative system (\Cref{fig:CAFEs_system}), where the clamping mechanism allows both independent and synchronized operation. Individual units can clamp on/off autonomously, while synchronized clamping ensures coordinated movement for collaborative tasks, maximizing workspace efficiency.

\section{EXPERIMENTAL SETUP}

\begin{figure}[h]
    \centering
    \includegraphics[width = 0.42\textwidth]{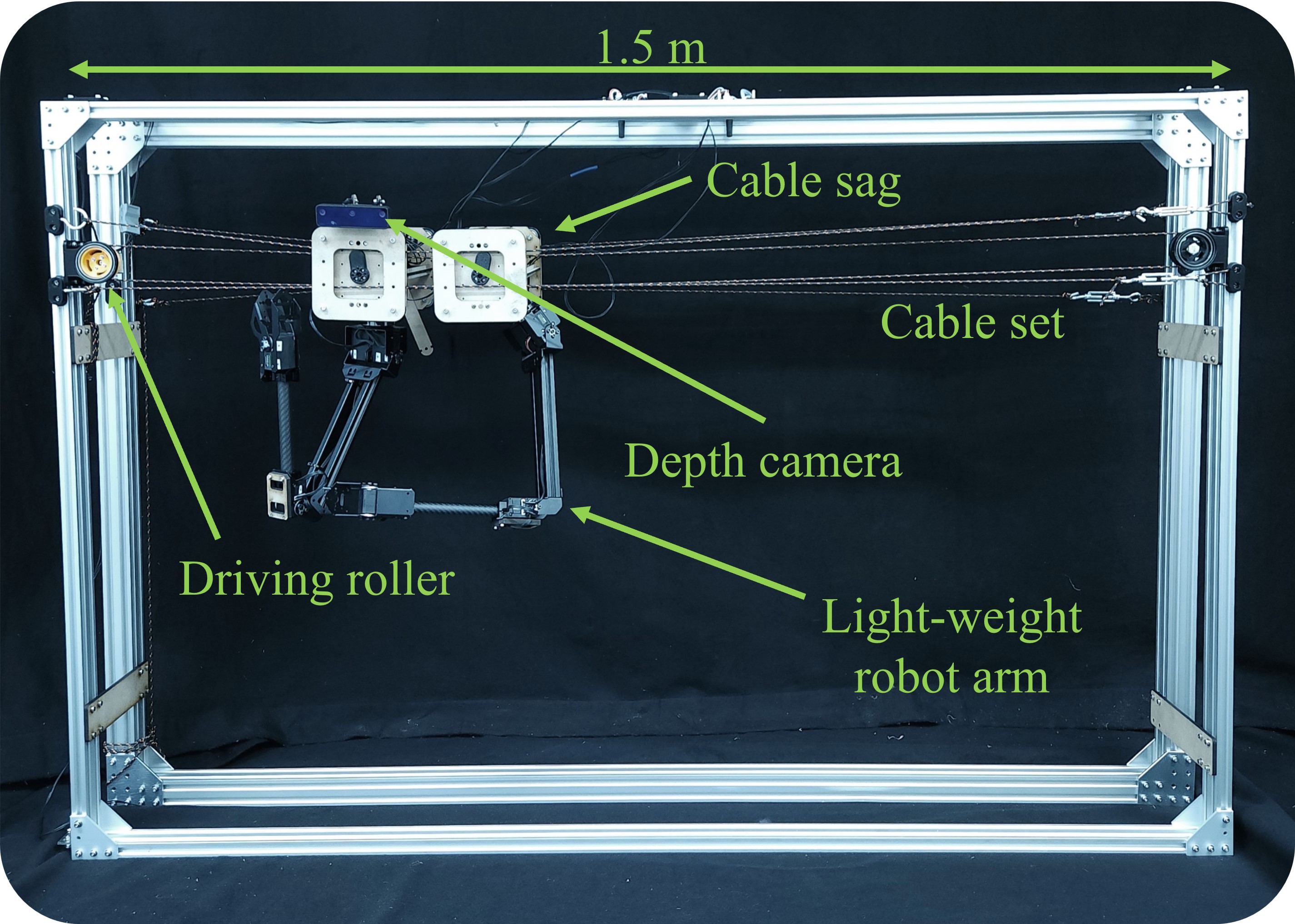}
    \caption{Experimental setup with motion capture}
    \label{fig:experiment_setup}
\end{figure}

To evaluate the proposed system, we set up a 1.5 m long cable mounted on a fixed aluminum structure, as illustrated in \Cref{fig:experiment_setup}. Two \emph{CAFEs} were deployed and controlled using MATLAB to follow various motions, as shown in \Cref{fig:motion_fig_set}, with motion data captured using the OptiTrack motion capture system.

The toggle motors (Dynamixel XL430-W250-T), with a stall torque of 1.4 Nm, a speed of 340°/s, a 90° switch angle, and a 0.3-second switching time, were used for the clamping joints. The rolling cable motor (Dynamixel XL430-W250-T) had a stall torque of 1.4 Nm and a movement speed of 120 mm/s. The stiffness of the 3 mm-diameter cables was measured at 18,148.5 N/m, with six cables under approximately 60 kg of pretension each, as shown in \Cref{fig:overview_ccafe}. The pretension was adjusted using screws, and the value was estimated by calculating the cable angles relative to the horizontal line with a known mass hanging in the middle.
A 2 mm-thick thermoplastic elastomer was used as the deformable material in the clamping mechanism (\Cref{fig:shematic_clamping}).
Each \emph{CAFE} weighed 1.4 kg, with the robotic arm weighing 0.9 kg, and a movable range of 1.3 m. The remaining components were constructed from laser-cut wood and 3D-printed materials.

\section{EXPERIMENT RESULTS}

This section presents the results of experiments assessing the performance of \emph{CAFEs}, focusing on the mechanism's characteristics, accuracy, precision, repeatability, spring-mass model, and scalability. A final demonstration is also given for some agricultural inspired tasks.

\begin{figure}[!htb]
    \centering
    \vspace{0.2cm}
    \begin{minipage}[b]{0.45\textwidth}
        \centering
        \includegraphics[width=\textwidth]{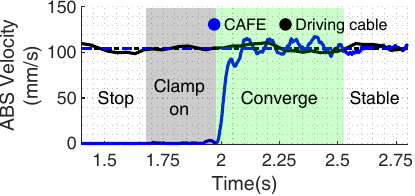}
        \centerline{(a) Clamp on}
        \label{fig:estimate_position_and_theory_position_motive_49}
    \end{minipage}
    \begin{minipage}[b]{0.45\textwidth}
        \centering
        \includegraphics[width=\textwidth]{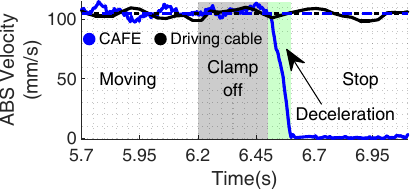}
        \centerline{(b) Clamp off}
        \label{fig:stop_converge}
    \end{minipage}
    \begin{minipage}[b]{0.45\textwidth}
        \centering
        \includegraphics[width=\textwidth]{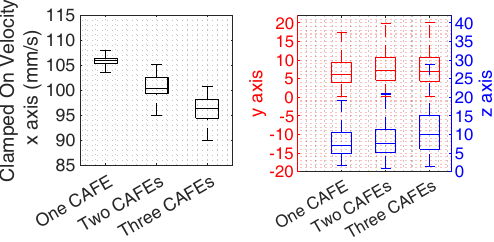}
        \centerline{(c) Velocity variation on $x$-$y$-$z$ axis}
        \label{fig:plate_cable_vel}
    \end{minipage}
    \caption{Motion captured result for \Cref{fig:motion_1}: Time for one CAFE to clamp on/off (grey area) and converge to the target velocity (green area)}
    \label{fig:sub_fig_set1}
\end{figure}

\subsection{Clamping and switching characteristic of the mechanism}

To explore the dynamic changes during state transitions and the influence of one \emph{CAFE} on another, the motions shown in \Cref{fig:motion_1} were executed under varying loads. The roller provided continuous motion at 105 mm/s, controlled by a proportional gain controller. Initially, different numbers of \emph{CAFEs} remained stationary, then transitioned to a moving state, and finally stopped. As shown in \Cref{fig:stop_converge,fig:estimate_position_and_theory_position_motive_49}, the \emph{CAFE} could achieve transition between 0 and 100 mm/s instantly.

The sudden change in load induced an impulsive force on the driving cable leading to damping effects. Consequently, the system exhibited more vibrations and incurred higher position errors with a greater number of \emph{CAFEs} clamped on simultaneously, especially in vertical ($z$) direction.

\subsection{Precision of open loop motion}

\begin{figure}[ht]
\centering  
 \vspace{0.2cm}
 \includegraphics[width = 0.45\textwidth]{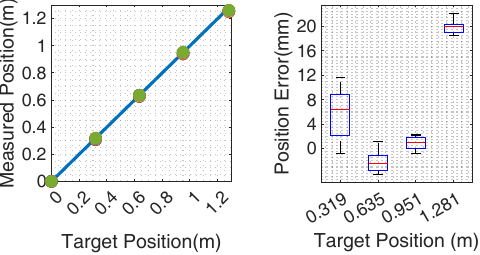}
    \caption{Motion captured result for \Cref{fig:motion_2}: Error of repeating go-and-stop motion for 1.25m long travel with one CAFE and target velocity $50 mm/s$}
    \label{fig:position_displacment_estimate_error}
\end{figure}
To evaluate the position accuracy and precision of the clamping robot, two sets of trajectories were conducted, as depicted in \Cref{fig:motion_2,fig:motion_3}. First, the platform aimed to reach predefined target positions spanning $0.2 m$ to $1.3 m$ using open-loop control. The desired position was estimated based on the roller's velocity and measured by the Optitrack system. As shown in \Cref{fig:position_displacment_estimate_error}, position errors accumulated over longer travels. However, the total position error after the $1.25 m$ traversal was within an acceptable range remained under $25 mm$, which equates to less than $2\%$.  

\begin{figure}[h]
    \centering
    \includegraphics[width=0.95\linewidth]{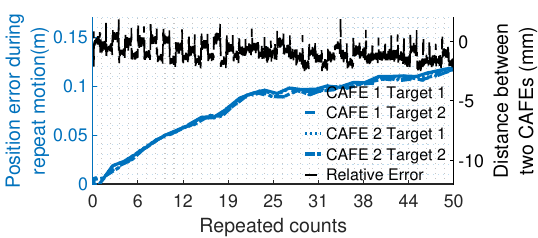}
    \caption{Motion captured result for \Cref{fig:motion_5}: Evaluation of the relative displacement of two \emph{CAFEs} for repeated synchronous motion}
    \label{fig:sync_motion_travel_with_error}
\end{figure}

To further assess the platform's precision, a repeating motion was executed with two platforms synchronously clamping on and off the driven cable, between two positions for 50 repeats, with a fixed cable speed of $100 mm/s$. This motion sequence involved moving forward, stopping, and then moving backward and stopping.
As shown in \Cref{fig:sync_motion_travel_with_error}, the error between the target and the measured position accumulated with more repeated motions. However, the relative position error of two CAFEs remains less than $3$ mm only, highlighting the advantages of the robot design for task that require synchronous movement.

The results demonstrated that the system achieved significant accuracy and precision, even without close-loop control. However, some accumulated error may arise from mechanical components like cable knots and pulleys, as changes in friction and tension occur when the knot passes through the roller. 

\begin{figure}[h]
    \centering   
    \vspace{0.2cm}\includegraphics[width=0.95\linewidth]{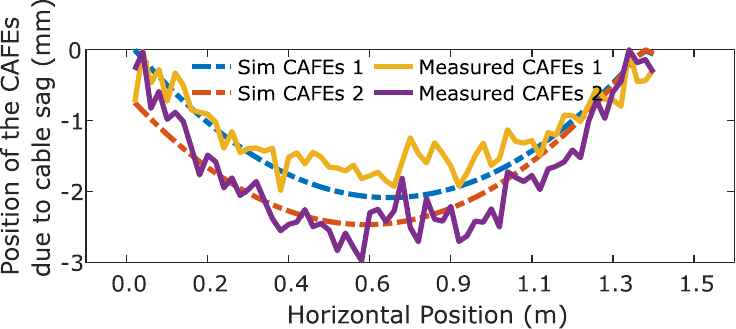}
    \caption{Comparison between the simulated model and the average vertical displacement of two CAFEs, based on the motion captured in \Cref{fig:motion_5}}
    \label{fig:sim_vs_measured}
\end{figure}
\subsection{Tension Distribution and System Scalability}

To explore the system's scalability, the cable sag and the number of \emph{CAFEs} per cable length were estimated using the dynamic model presented in \Cref{sec_Ten}. The model was first validated by comparing it with measurement data to ensure accuracy. The experiment setup was experienced the 60kg pretension per cable and resulted in a 2.5mm cable sag due to the close placements of the two \emph{CAFEs}. As shown in \Cref{fig:sim_vs_measured}, the estimated cable sag closely matched the measured data from \Cref{fig:motion_5}. The simulation accurately predicted cable sag during motion, particularly in the vertical displacement of the two \emph{CAFEs}. Moreoever, since the \emph{CAFEs} were positioned close to each other, the relative sag error was minimal and did not impact the cooperative tasks.

\begin{figure}[h]
    \centering    \includegraphics[width=1.0\linewidth]{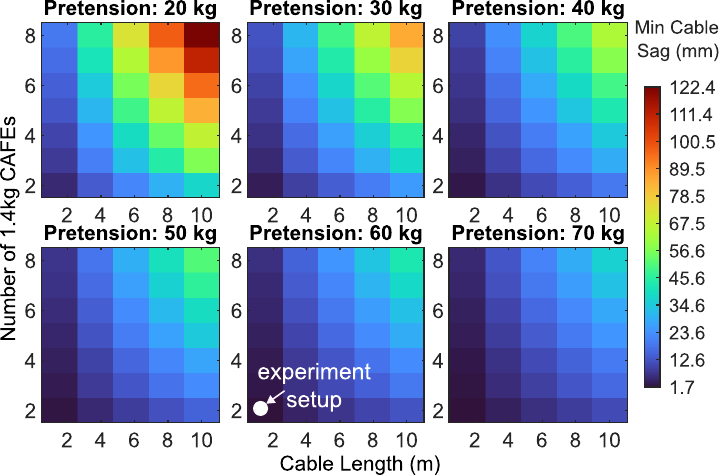}
    \caption{Scalability of the 1.4 kg robot regarding the pretension, cable sag, cable length and number of evenly deployed CAFEs. More CAFEs or longer cable requires higher pretension to maintain low cable sag}
    \label{fig:cable_sag_pretension_distget}
\end{figure}

The dynamic model was then applied to estimate cable sag, cable pretension, and the possible number of deployed \emph{CAFEs} per cable length for large-scale operations. The results, shown in \Cref{fig:cable_sag_pretension_distget}, indicate that for a 10 m cable span, the cable sag for two evenly distributed robots is 13 mm. This suggests that the most sagged robotic arm would need to compensate for a 13 mm vertical displacement to reach the zero horizontal line (the height of the cable anchor). Using the proposed method, application requirements can be estimated based on the known number of robots, robot's mass, cable length, pretension, and cable stiffness.
Additionally, the simplified dynamic model could be use to predict the oscillation mode and cancellation in future developments.

\begin{figure}[ht]
    \centering  \vspace{0.1cm}   \includegraphics[width=0.96\linewidth]{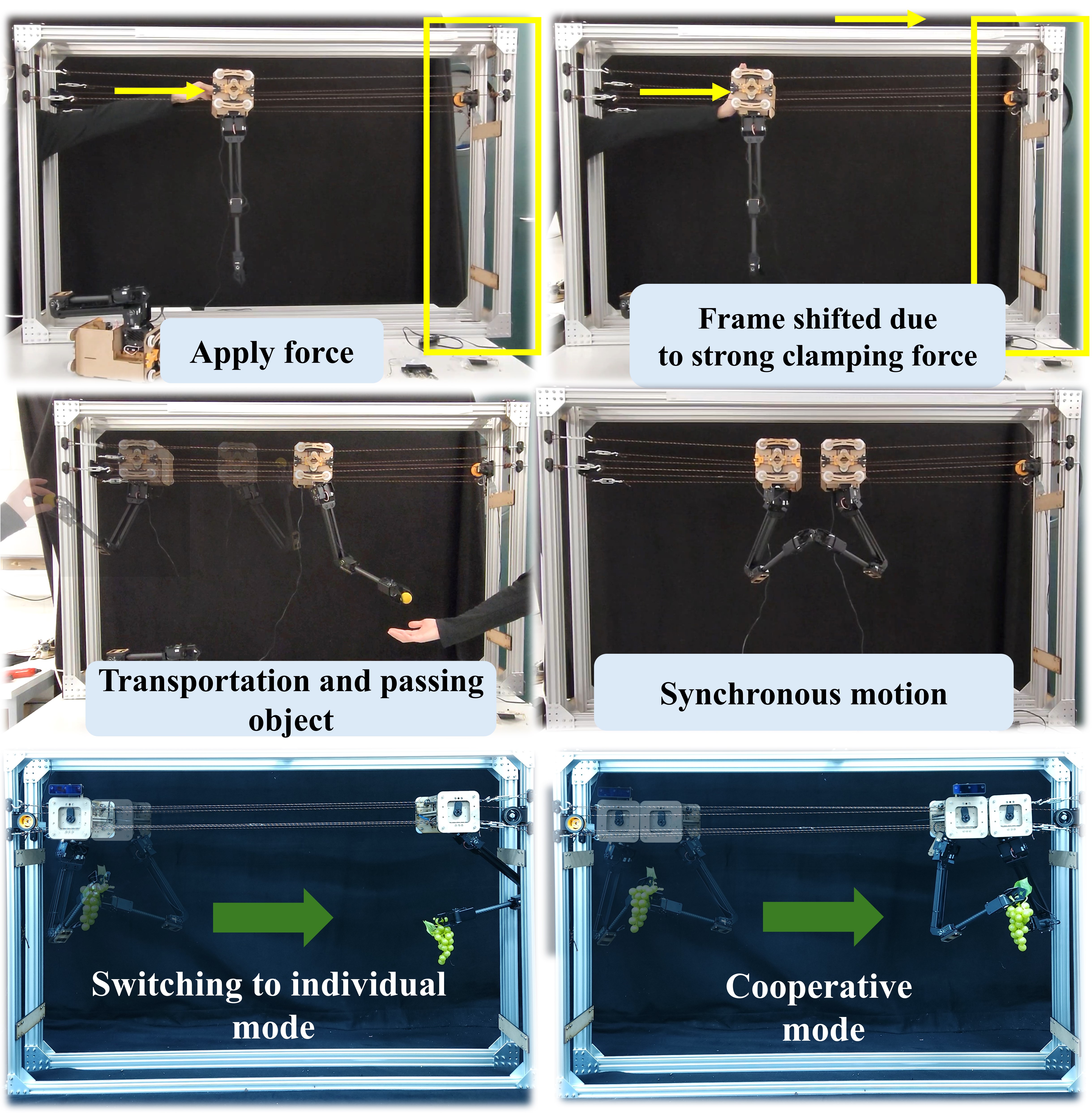}
    \caption{Clamping force test and examples of motion type in agriculture tasks}
    \label{fig:exp_all_hori}
\end{figure}
\subsection{Demonstration of the System}

To demonstrate the collaborative tasks between multiple \emph{CAFEs}, two prototypes were tasked with performing simple transition and synchronous motions, as shown in \Cref{fig:exp_all_hori}. The experiment aimed to simulate various operation modes, as depicted in  \Cref{fig:CAFEs_system}. The results indicated that stable and accurate positioning of the floating platforms was crucial, especially during cooperative transitions. Moreover, since the \emph{CAFEs} were positioned close to each other, the clamping force was tested through demonstrations of transportation and object-passing tasks.

\section{CONCLUSIONS}
The development and testing of the cable-driven Collaborative Agricultural Floating End-effectors (CAFEs) system demonstrate its potential for large-scale agricultural applications by integrating collaborative capabilities, a fast-clamping mechanism, and dynamic scalability, overcoming the limitations of traditional cable-driven robots restricted to a single end-effector. Experiments confirmed high positioning accuracy, stable cooperative motion, and repeatability, while the validated dynamic model for tension distribution and cable sag compensation ensures stability in larger deployments. The results highlight the feasibility of CAFEs for harvesting, monitoring, and object transport, improving operational efficiency. Future work will focus on enhancing control algorithms, optimizing hardware, and refining motion planning to address multi-robot coordination, while further development of anti-oscillation damping will improve system stability, positioning CAFEs as a key technology in agricultural automation.

\bibliographystyle{IEEEtran}
\bibliography{ref.bib}

\end{document}